\documentclass[journal,a4paper]{IEEEtran}
\IEEEoverridecommandlockouts
\usepackage{cite}
\usepackage{amsmath,amssymb,amsfonts}
\usepackage{algorithmic}
\usepackage{textcomp}
\usepackage{xcolor}
\usepackage{amsmath}
\usepackage[export]{adjustbox}
\usepackage{float}
\usepackage{graphicx}
\usepackage{multirow}
\usepackage[]{siunitx}
\DeclareSIUnit{\degree}{°}
\DeclareSIUnit{\deg}{deg}
\DeclareSIUnit{\nothing}{\relax}
\DeclareSIUnit{\pixel}{px}

\def\BibTeX{{\rm B\kern-.05em{\sc i\kern-.025em b}\kern-.08em
    T\kern-.1667em\lower.7ex\hbox{E}\kern-.125emX}}
\begin{document}

\title{Multi-sensory Anti-collision Design for \\  
Autonomous Nano-swarm Exploration}





\pagenumbering{gobble}
\markboth{This paper has been accepted for publication in the IEEE 30th International Conference on Electronics, Circuits and Systems (ICECS).}{}



\author{
\IEEEauthorblockN{
Mahyar Pourjabar\IEEEauthorrefmark{1},
Manuele Rusci\IEEEauthorrefmark{2},
Luca Bompani\IEEEauthorrefmark{1},
}
\IEEEauthorblockN{
Lorenzo Lamberti\IEEEauthorrefmark{1},
Vlad Niculescu\IEEEauthorrefmark{3},
Daniele Palossi\IEEEauthorrefmark{3}\IEEEauthorrefmark{4},
and Luca Benini\IEEEauthorrefmark{1}\IEEEauthorrefmark{3}
}

\IEEEauthorblockA{\IEEEauthorrefmark{1} Department of Electrical, Electronic and Information Engineering (DEI), University of Bologna, Italy}
\IEEEauthorblockA{\IEEEauthorrefmark{2} Department of Electrical Engineering, KU Leuven, Belgium}
\IEEEauthorblockA{\IEEEauthorrefmark{3} Integrated Systems Laboratory (IIS), ETH Z\"urich, Switzerland}
\IEEEauthorblockA{\IEEEauthorrefmark{4} Dalle Molle Institute for Artificial Intelligence (IDSIA), USI-SUPSI, Switzerland}
Contact author: mahyar.pourjabar2@unibo.it
}

\maketitle

\begin{abstract}
This work presents a multi-sensory anti-collision system design to achieve robust autonomous exploration capabilities for a swarm of \SI{10}{\centi\meter}-side nano-drones operating on object detection missions. 
We combine lightweight single-beam laser ranging to avoid proximity collisions with a long-range vision-based obstacle avoidance deep learning model (i.e., PULP-Dronet) and an ultra-wide-band (UWB) based ranging module to prevent intra-swarm collisions. 
An in-field study shows that our multi-sensory approach can prevent collisions with static obstacles, improving the mission success rate from 20\% to 80\% in cluttered environments w.r.t. a State-of-the-Art (SoA) baseline. 
At the same time, the UWB-based sub-system shows a 92.8\% success rate in preventing collisions between drones of a four-agent fleet within a safety distance of $\sim$\SI{65}{\centi\meter}. 
On a SoA robotic platform extended by a GAP8 multi-core processor, the PULP-Dronet runs interleaved with an objected detection task, which constraints its execution at 1.6~frame/s.
This throughput is sufficient for avoiding obstacles with a probability of $\sim$40\% but shows a need for more capable processors for the next-generation nano-drone swarms.
\end{abstract}


\section{Introduction}\label{sec:intro}

Thanks to their broad applicability, swarms of nano-sized autonomous unmanned aerial vehicles (nano-UAVs or nano-drones), i.e., flying robots with sub-\SI{10}{\centi\meter} diameter, are gaining momentum in academia and industry~\cite{Awasthi2022UAVsFI}.
Swarms of nano-drones can act in rescue missions and hazard areas, exploring cluttered indoor environments and narrow spaces~\cite{barcis2022self}.
They can also safely operate near humans and interact with them thanks to their reduced form factor~\cite{cereda2021improving}.

Two fundamental functionalities needed by a nano-UAV swarm are the capability to prevent collisions, both with objects in the environment (i.e., obstacle avoidance) and between the swarm's agents (i.e., intra-swarm collision avoidance) and detect objects/subjects of interest (i.e., object detection).
However, this multi-tasking scenario is challenged by the limited resources available aboard nano-UAVs, which offer ultra-constrained payload ($\sim$10 grams) and power envelope (a few Watts in total, of which a few 100s \SI{}{\milli\watt} for the onboard electronics).
Therefore, nano-UAVs come with minimal low-bandwidth memories (sub-\SI{512}{\kilo\byte} on-chip), limited processing power (a few 100s \SI{}{\mega Op/\second}), and sensors (e.g., low-resolution cameras)~\cite{DCOSS19}.

To cope with these limitations, State-of-the-Art (SoA) nano-drones take advantage of bio-inspired lightweight algorithms~\cite{McGuire2018ACS, Lamberti2023BioinspiredAE} for exploring the environment and simple Time-of-Flight (ToF) distance sensors~\cite{mcguire2019minimal, Duisterhof2021SniffyBA, Lamberti2023BioinspiredAE} or vision-based pipelines~\cite{DCOSS19, Lamberti2022TinyPULPDronetsSN} to prevent collision with obstacles.
Additionally, to avoid intra-swarm collisions, many recent works leverage ultra-wide-band (UWB) radio technology, which can be used for ranging and data exchange~\cite{Duisterhof2021SniffyBA, Li2020AnAS}.
Finally, to address the object detection, vision-based perception is crucial, as shown in~\cite{Lamberti2023BioinspiredAE}, which exploits a computationally intensive single shot detector (SSD) algorithm running at \SI{1.7}{\hertz} on a nano-drone.

This work addresses such a challenging scenario by proposing a novel nano-drone swarm onboard system targeting a reliable autonomous exploration and vision-based object detection on a highly resource-constrained robotic platform.
We focus on improving the collision avoidance ability within the exploration task in an unknown and obstacle-populated indoor area by combining the advantages of front-looking QVGA images and four single-beam ToF sensors (front, back, left, right).
Monochrome low-resolution images can be noisy, but they have a large field of view, and they provide semantic information that we process with the open-source PULP-Dronet convolutional neural network (CNN)~\cite{Lamberti2022TinyPULPDronetsSN} which predicts probabilities of collision.
ToF sensors provide accurate distance measurements at close distances ($<$\SI{4}{\meter}), which is complementary to the camera.
Then, the UWB-based ranging measurements give the swarm additional, reliable information, which we use to maintain a minimal safe distance between drones, preventing intra-swarm collisions.

This paper makes the following contributions:
\begin{itemize}
    \item we propose a compact multi-sensory design strategy for nano-drone swarms, enabling object detection and collision avoidance encompassing a camera, four ToF sensors, and an UWB radio;
    \item we empirically evaluate the effectiveness of our multi-sensory approach (full deployment) in various field settings, conducting multiple rounds of experimentation.
\end{itemize}

Our study shows that the obstacle avoidance capability of our nano-drone swarm increases by +60\% within cluttered environments w.r.t. a third-party~\cite{Duisterhof2021SniffyBA} baseline system, which employs only ToF sensors. 
At the same time, the intra-swarm anticollision performance reaches 92.8\% when tested on a swarm of four drones. 
Conversely, our closed-loop performance analysis also shows the current limitation of our system, which is the available onboard computing power.
Interleaving the PULP-Dronet CNN with the SSD algorithm on the GAP8 multi-core System-on-Chip (SoC) bounds the final CNN throughput to \SI{1.6}{\hertz}.
This performance is 5$\times$ less than required in our field tests for minimum throughput for a probability $>$80\% of crash avoidance at a mean flight speed of \SI{0.5}{\meter/\second}.
\begin{figure}[t]
    \centering
    \includegraphics[width=0.9\columnwidth]{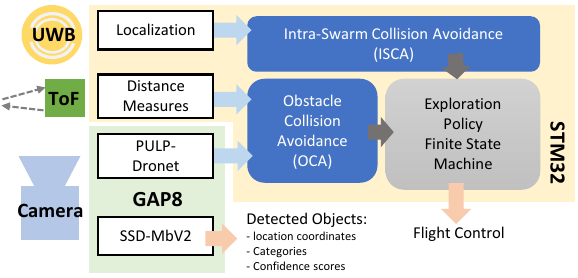}
    \caption{Multi-sensory design approach for autonomous navigation and object search on a swarm of nano-drones.}
    \label{fig:intergrated-system-overview}
\end{figure}

\section{Background}

Our work is based on the reference platform proposed by \textit{Lamberti et al.}~\cite{Lamberti2023BioinspiredAE}, which is built around a commercial off-the-shelf (COTS) Crazyflie 2.1 quadrotor, a nano-drone with a weight of \SI{27}{\gram} and a diameter of \SI{10}{\centi\meter}. 
The platform, which is powered by an STM32F405 MCU (referenced as STM32 in this paper), features multiple COTS sensors, available in the form of pluggable \textit{decks}: 
\textit{(i)} a \textit{Flow deck} for optical flow and height measurements, 
\textit{(ii)} a \textit{Multi-ranger deck} with 5 VL53L1x laser-ranging sensors and  
\textit{(iii)} an \textit{AI-deck} hosting a low-power QVGA grayscale camera (Himax HM01B0). 
The sensors of the Multi-ranger deck compute the distance with respect to the facing objects using a ToF measurement. 
When the Multi-ranger deck is mounted on the Crazyflie, four sensors are used to monitor the lateral sides of the drone. 
The AI-deck also includes an 8-core MCU, GAP8 \cite{flamand2018gap}, working as on-board visual processing engine. 
Additionally, we augment the nano-agent design with an UWB radio module, i.e., the \textit{Loco-Positioning deck\footnote{https://www.bitcraze.io/products/loco-positioning-deck/}}, serving the intra-swarm localization and anti-collision.

The reference design relies on a bio-inspired autonomous navigation policy fed by the ToF sensor measurements. 
The drone flies following an obstacle-free direction until one of the front, left, or right ToF sensors detects an object at a distance of less than \SI{1}{\meter}. 
To prevent collisions, the agent reduces the velocity to zero and rotates by a random angle. 
During the exploration, the drone performs a visual object search task: a deep learning-based SSD object detector with a MobileNetV2 backbone~\cite{liu2016ssd},  i.e., SSD-MbV2, runs on the GAP8 MCU to extract the locations, categories and confidence scores of any target object present in the recorded images.

Conversely, the PULP-Dronet~\cite{Lamberti2022TinyPULPDronetsSN} CNN enables vision-based autonomous navigation aboard nano-drones.
This model predicts two outputs: a collision probability and a steering angle.
Starting from PULP-Dronet, we leverage its obstacle avoidance capability in cluttered environments.
Given the strict memory and real-time constraints of our UAV, we adopt a lightweight variant of PULP-Dronet, i.e., Tiny-PULP-Dronet~\cite{Lamberti2022TinyPULPDronetsSN}, featuring a half number of channels in each convolution layers w.r.t. the original version~\cite{DCOSS19}.
The Tiny-PULP-Dronet, with a total number of  \SI{83.9}{\kilo\nothing} parameters, reaches a peak throughput of \SI{62.9}{frame/\second} on GAP8.

\begin{figure}[t]
    \centering
    \includegraphics[width=\columnwidth]{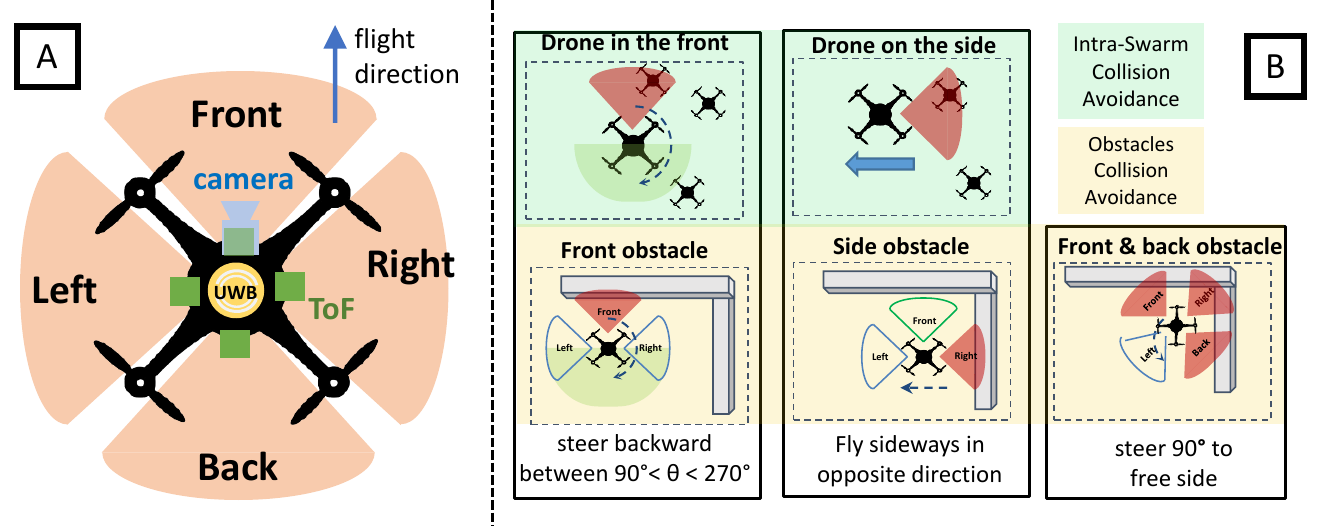} 
    \caption{[A] System setup [B] Exploration policy based on the multi-sensory collision avoidance logic.}
    \label{fig:swarm_autonomous_navigation_policy_flowchart}
\end{figure}

\section{System Design} \label{System Design}

Fig.~\ref{fig:intergrated-system-overview} illustrates the proposed multi-sensory design to enable nano-swarm autonomous exploration and object search in a cluttered environment.
Every agent uses data from the UWB, ToF, and camera sensors to detect potential collisions with obstacles or other nano-drones.
More in detail, a \textit{Obstacle Collision Avoidance} (OCA) module takes the ToF distance measures and the probability of collision scores computed by PULP-Dronet to assess the presence of nearby obstacles. 
Concurrently, the \textit{Intra-Swarm Collision Avoidance} (ISCA) task reports if any other drone of the fleet is within a critical distance. 
The OCA and ISCA modules, detailed below, feed an Exploration Policy Finite State Machine that drives the exploration in the unknown environment.  
Lastly, the visual object detection task, i.e., SSD-MbV2 inference, runs on the GAP8 accelerator to detect the objects of interest. 

For the purpose of collision avoidance, we identify four areas in the drone's horizontal plane: the \textit{front}, \textit{left}, \textit{right}, and \textit{back} zones (Figure~\ref{fig:swarm_autonomous_navigation_policy_flowchart}-A). 
Each zone is monitored by one of the ToF sensors; the onboard camera records images in front of the drone. 
The exploration policy commands a straight trajectory until an obstacle or another nano-drone is revealed in one of the four surrounding zones. 
In this case, the navigation stops, and the drone reacts according to the strategy visualized in Fig.~\ref{fig:swarm_autonomous_navigation_policy_flowchart}-B. 
When an object or another agent is detected in the front zone, the finite state-machine controls a random rotation between 90\textdegree and 270\textdegree.
If both backward and front zones are occupied, a rotation of 90\textdegree is forced to avoid trapping between the obstacles.
In case of right or left zones' occupancy, the drone moves to the opposite side to ensure a safe distance.

\subsection{Obstacles Collision Avoidance (OCA)}

The OCA module relies on the measurements of the ToF sensors, sampled at \SI{20}{\Hz}, and the collision probability predictions of the PULP-Dronet. 
A risk of collision in one of the four zones is assessed if five consecutive ToF measurements from the corresponding sensor are below a critical threshold of~\SI{1}{\meter}~\cite{Lamberti2023BioinspiredAE}.
This processing task, consisting only of a comparison, constitutes a negligible workload on the STM32.
Conversely, a collision risk in the front zone is assessed if PULP-Dronet returns a collision probability score higher than 0.7 on two consecutive frames.
In our system, PULP-Dronet runs on the GAP8 MCU, alongside the SSD-MbV2 object detection algorithm. 
Because both algorithms share the same compute resource, we adopt an interleaved execution model. 
Hence, the maximum detection rate of the PULP-Dronet is limited by the throughput of the SSD-MbV2 execution.

\subsection{Intra-Swarm Collision Avoidance (ISCA)}

By leveraging the UWB technology, every drone localizes itself over time, i.e., computes its coordinates according to a relative reference frame. 
The localization coordinates are then shared in the swarm using UWB broadcast communications, enabling the ISCA module to determine if an agent is at a critical distance in one of the four surrounding zones. 

The localization is collaboratively performed by the agents using UWB ranging.
A ranging operation involves an initiator and a responder UWB modules. 
The first starts ranging operations by transmitting a message, while the second stays in listening mode.
After defining a global ranging schedule, one of the drones becomes the initiator and performs ranging with every other component of the swarm in turn. 
Once the round is completed, the next scheduled drone takes the initiator role and drives the ranging process. 
The UWB ranging measurements feed a Kalman filter that tracks the location over time. 
This filter also incorporates data from the optical flow sensor and the onboard accelerometer.
In our design, the initial take-off position of every drone in the global frame is known. 

To design the UWB localization function, which runs on the STM32 MCU, we rely on the UWB Software Library proposed in~\cite{Pourjabar2023LandL}. 
Differently from this previous work, we dynamically switch the drone roles between initiator and receiver. 
Hence every agent can coordinate the UWB procedure without the infrastructural cost of fixed initator nodes, i.e., the anchors~\cite{Hyun2019UWBbasedIL}.
\section{Experimental results}

\begin{figure}[t]
    \centering
    \includegraphics[width=1\columnwidth]{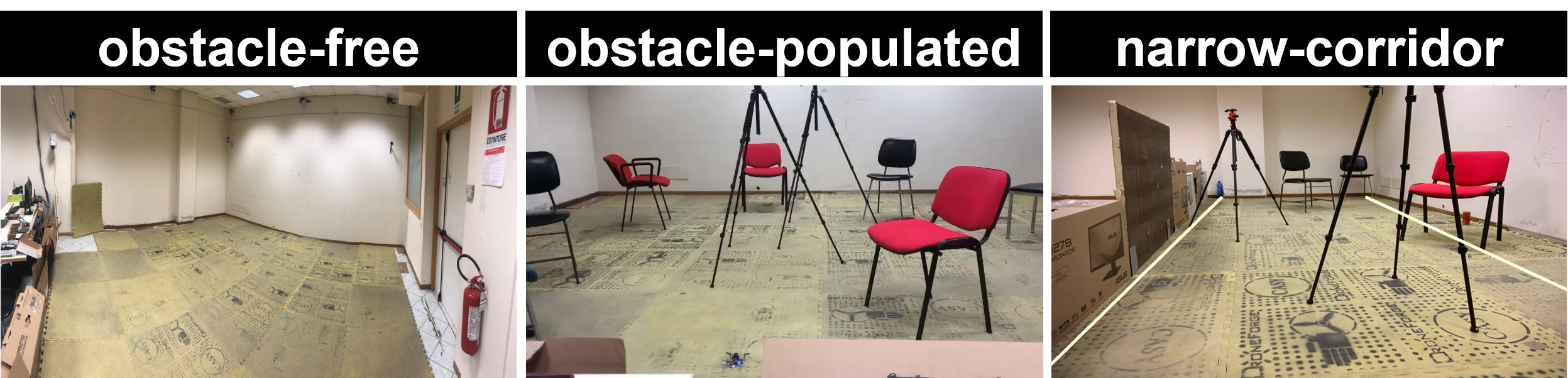} 
    \caption{Different environment setting for the in-field assessment. The test arena is 6.6x\SI{5.6}{\meter}$^2$ wide, restricted to 1.95x\SI{4.5}{\meter}$^2$ for the narrow-corridor scenario.}
    \label{fig:environment_setup}
\end{figure}

\subsection{In-field Experiments}

\textbf{Experiment 1}.
\textit{Visual detection improves the collision success rate in cluttered environments from 20\% to 80\%}.

We test our OCA method, namely ToF\&PULP-Dronet, on a single drone setup to assess its effectiveness with respect to systems using a ToF-only collision avoidance method.
Fig.~\ref{fig:environment_setup} illustrates the three considered test environments: (A) \textit{obstacle-free}, (B) \textit{obstacle-populated} and (C) \textit{narrow-corridor}.
In environments B and C, we challenge the nano-drone autonomous explorations using narrow obstacles, i.e., chair legs and tripods. 
The latter scenario differentiates because the obstacle positioning originates a corridor for the drone flight.

In this experiment, a nano-drone takes off from a known position and explores the unknown environment driven by the policy described in Section~\ref{System Design}. 
For both the ToF-only and ToF\&PULP-Dronet settings, we perform 5 and 10 runs, respectively, for the obstacle-free and narrow corridor environments, while 20 runs are experimented in the obstacle-populated environment. 
During all the tests, the flight velocity is set to~\SI{0.5}{\meter/\second}, and the drones fly at~\SI{0.3}{\meter} altitude. 
PULP-Dronet runs at a throughput of~\SI{5}{\Hz}, lower than the model peak throughput~\cite{Lamberti2022TinyPULPDronetsSN} because of the extra operations, such as image capturing time, image compression, and logging.
We track the drone trajectories using an external motion capture system, and we manually annotate the crashes with the obstacles.

\begin{table}[t]
\centering
\caption{In-field experiments results on the assessment of obstacle avoidance performance.}
\label{table:ToF_dronet_advantage}
\resizebox{\linewidth}{!}{
\begin{tabular}{clccc}
\hline
Env. & Parameters & SniffyBug & ToF-only & \begin{tabular}[c]{@{}c@{}}ToF\&\\ PULP-Dronet\end{tabular} \\ \hline
\multicolumn{1}{c|}{\multirow{3}{*}{\begin{tabular}[c]{@{}c@{}}obstacle\\ free\end{tabular}}} & Crash-free rounds & 5 /5 & 5 /5 & - \\
\multicolumn{1}{c|}{} & Avg. crash/min & 0 & 0 & - \\
\multicolumn{1}{c|}{} & Avg. coverage/min{[}\%{]} & 20.7 & 22.4 & - \\ \hline
\multicolumn{1}{c|}{\multirow{3}{*}{\begin{tabular}[c]{@{}c@{}}obstacle\\ populated\end{tabular}}} & Crash-free rounds & 4 /20 & 3 /20 & 16 /20 \\
\multicolumn{1}{c|}{} & Avg. crash/min & 1.8 & 1.5 & 0.2 \\
\multicolumn{1}{c|}{} & Avg. coverage/min{[}\%{]} & 33.2 & 22.6 & 10.3 \\ \hline
\multicolumn{1}{c|}{\multirow{3}{*}{\begin{tabular}[c]{@{}c@{}}narrow\\ corridor\end{tabular}}} & Crash-free rounds & 0 /10 & 1 /10 & 7 /10 \\
\multicolumn{1}{c|}{} & Avg. crash/min & 2.6 & 2.4 & 0.4 \\
\multicolumn{1}{c|}{} & Avg. coverage/min{[}\%{]} & 19.8 & 45.2 & 26.7 \\ \hline
\end{tabular}
}
\end{table}

Table~\ref{table:ToF_dronet_advantage} shows the results of our in-field experiments collected during 100 testing runs and reports the number of collision-free missions and the average number of collisions per minute of flight. 
We also list the average coverage area per minute (expressed in \% of the total area), accounted as the number of visited cells w.r.t. the total number of cells (a cell is an area of 5x\SI{5}{\centi\meter}$^2$).
In addition to our ToF-only and ToF\&PULP-Dronet systems, we benchmark the anti-collision capability of a single drone of the SniffyBug swarm~\cite{Duisterhof2021SniffyBA} solution, which also relies on the ToF information.

In the obstacle-free environment, the ToF solutions already show perfect performance (100\% collision-free). 
Hence, we did not run any experiments using PULP-Dronet. 
Conversely, for the obstacle-populated environment, only 20\% or 15\% of the trials are successful for, respectively, the SniffyBug and our ToF-only system, dropping to 0\% in the narrow corridor for both systems. 
The poor results are motivated by the missed detection of the narrow obstacles by the ToF sensors. 
The proposed ToF\&PULP-Dronet achieves instead a success rate of 80\% and 70\% in the two environments. 
In the narrow-corridor setup, the ToF-only system shows a coverage-per-minute of 45.2\%, which is higher than Sniffy-bug and ToF\&PULP-Dronet solutions.
The latter, in particular, suffers from detecting the corridor with a potential risk of collision.

\begin{figure}[b]
    \centering
    \includegraphics[width=1.0\columnwidth]{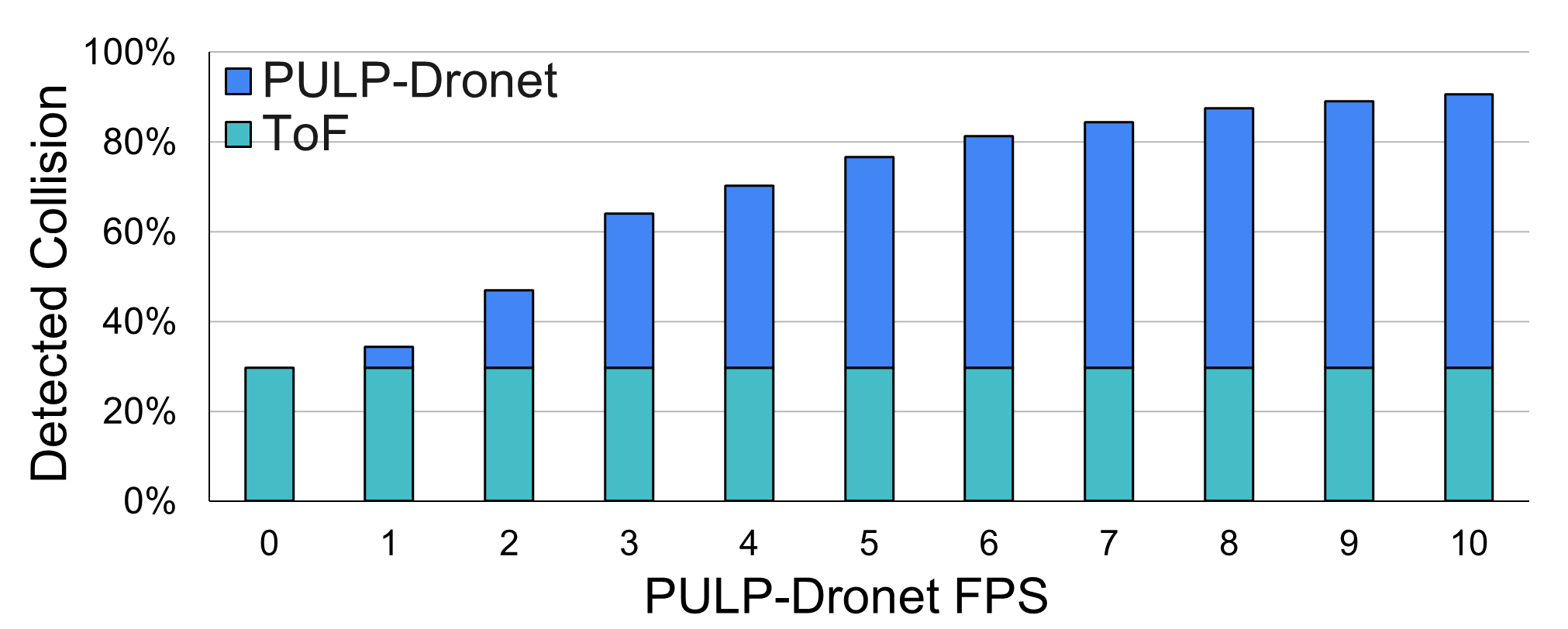} 
    \caption{PULP-Dronet obstacle detection at different throughput.}
    \label{fig:bar_chart_dronet_vs_ToF_detection_share}
\end{figure}

\textbf{Experiment 2}. \textit{A PULP-Dronet throughput of $\geq$\SI{8}{\Hz} is recommended at~\SI{0.5}{\meter/\second} speed for accurate collision avoidance}.

To derive the minimal throughput requirement of the visual pipeline, we analyze the obstacle detection ability of the PULP-Dronet for multiple execution rates. 
We record a video dataset of a nano-drone approaching 8 obstacles (6 chairs and 2 tripods) from a distance of \SI{60}{\centi\meter} at a speed of \SI{0.5}{\meter/\second}. 
In total, we collected 64 videos at 10 FPS with an average number of 15 frames and at least 12 frames per video. 
At the same time, the ToF measurements were registered.

To analyze the impact of multiple throughputs from~\SI{1}{\Hz} to~\SI{10}{\Hz}, we subsample the acquired video and we apply the PULP-Dronet algorithm.
The results in Fig.~\ref{fig:bar_chart_dronet_vs_ToF_detection_share} show the number of detected collisions over the dataset.
The bottom bars denote the collision assessed only by the ToF (i.e., PULP-Dronet at 0 FPS), which does not vary across the throughput. 
At $2 FPS$, the benefit of PULP-Dronet is bounded to +17\% improvement with respect to the ToF-only solution. 
From $3 FPS$ to $6 FPS$, the boost in detection rate increases from 34\% to 51\%, while the performance gain reduces at higher FPS.
From this study, we conclude that the PULP-Dronet throughput of 8 FPS provides an optimal yet effective obstacle detection of 87.5\%, which is only 3\% lower than at 10 FPS.
 
\textbf{Experiment 3}. \textit{UWB-based Intra-Swarm Collision Avoidance achieves a 92.8\% success rate in preventing collisions}.   

The experiment is conducted with four nano-drones with the COTS Loco-Positioning deck, exploring in the obstacle-free environment at a fixed velocity of 0.5~\si{\meter/\second}.  
The system logs a timestamp if the ISCA module detects another drone of the swarm at a critical distance, i.e., ~\SI{65}{\centi\meter}.
This safety distance was empirically proved to be a safe distance to prevent collisions between nano-drones.
We collect data from 5 runs (586 seconds of total flight time), and we use the motion tracking system to record the ground truth positions of the agents.
We observed the ISCA module to correctly detect 92.8\% of the collision events, with precision and recall scores of, respectively, 96.2\% and 92.8\%, denoting a reliable behavior of the task, yet not perfect.
We motivate the miss-detections given the UWB ranging and localization errors.

\begin{table}[t]
\caption{Embedding Costs for the proposed Multi-Sensory Design. Model Parameters are stored in the Flash Memory. Memory refers to the maximum usage of L2 memory.}
\label{tab:system_cost}
\resizebox{\linewidth}{!}{
\begin{tabular}{clcccc}
\hline
MCU & Task & \begin{tabular}[c]{@{}c@{}}Throughput\\ {[}Hz{]}\end{tabular} & \begin{tabular}[c]{@{}c@{}}Parameters\\ {[}MB{]}\end{tabular} & \begin{tabular}[c]{@{}c@{}}Memory\\ {[}KB{]}\end{tabular} & \begin{tabular}[c]{@{}c@{}}Power\\ {[}mW{]}\end{tabular} \\ \hline
\multicolumn{1}{c|}{\multirow{2}{*}{STM32}} & UWB & 20 & - & 0.4 & 320.1 \\
\multicolumn{1}{c|}{} & ToF & 20 & - & 0.1 & 286~\cite{Lamberti2023BioinspiredAE} \\ \hline
\multicolumn{1}{c|}{\multirow{3}{*}{GAP8}} & SSD & 1.7 & 4.67 & 250 & 134~\cite{Lamberti2023BioinspiredAE} \\
\multicolumn{1}{c|}{} & PULP-Dronet & 62.9 & 0.0832 & 200.4 & 100.9 \\
\multicolumn{1}{c|}{} & SSD\&PULP-Dronet & 1.6 & 4.7532 & 250 & 133.1 \\ \hline
\end{tabular}
}
\end{table}

\subsection{System Integration Costs}

Table~\ref{tab:system_cost} reports the latency, memory, and power costs of the processing tasks running on the targeted robotic platform. 
On the STM32 MCU, the exploration policy executes every 20ms to ensure a prompt analysis of the ToF, UWB data, or the PULP-Dronet score.
The workload and memory costs of ToF and UWB processing tasks, both scheduled with a frequency of ~\SI{20}{\Hz}, are negligible, and the sensor power costs amount respectively to \SI{286}{\mW} and \SI{320.1}{\mW}, for a ranging process with 4 nano-drones.
On the AI-deck side, instead, the SSD-MbV2 execution constitutes the bottleneck for the multi-task integration. 
Because of the total \SI{589.2}{\ms} to run object detection, the PULP-Dronet, which takes only~\SI{15.9}{\ms} for the inference, can achieve an overall throughput of ~\SI{1.6}{\Hz}, which impacts the benefits of obstacle avoidance capabilities (less than +17\%).
\section{Discussion and Conclusion}

In this work, we present a lightweight and compact multi-sensory design for nano-swarm, operating a more reliable autonomous swarm exploration w.r.t. baseline system~\cite{Duisterhof2021SniffyBA}, along with executing object detection tasks on an onboard resource-limited processor. 
Thanks to the multi-modal obstacle detection via fusing PULP-Dronet probability and ToF-based sensing, collision avoidance can improve by +60\% in cluttered environments, while UWB-based intra-swarm anti-collision achieves an accuracy of 92.8\% in detecting critical events. 
When analyzing the system costs, we concluded that the state-of-the-art hardware i.e., GAP8, can not guarantee the highest obstacle avoidance capability of the proposed system because of the lack of computation power, calling for a new class of digital processors for this multi-sensory design. 
\section{Acknowledgement}\label{sec:acknowledgement}

This work has been partially funded by the Autonomous Robotics Research Center (ARRC) of the UAE Technology Innovation Institute (TII).
We thank the Center for Research on Complex Automated Systems and, in particular, Andrea Testa and Lorenzo Pichierri for their support.

\bstctlcite{IEEEexample:BSTcontrol}

\bibliographystyle{IEEEtran}
\bibliography{IEEEabrv,bibliography}

\end{document}